\documentclass[conference]{IEEEtran}
\IEEEoverridecommandlockouts
\usepackage{cite}
\usepackage{amsmath,amsfonts}
\usepackage{amssymb}
\usepackage{physics}
\usepackage{algorithm}
\usepackage{algpseudocode}
\usepackage{multirow}
\usepackage{graphicx}
\usepackage{textcomp}
\usepackage{xcolor}

\def\BibTeX{{\rm B\kern-.05em{\sc i\kern-.025em b}\kern-.08em
    T\kern-.1667em\lower.7ex\hbox{E}\kern-.125emX}}
\begin{document}

\title{Secure Navigation using Landmark-based Localization in a GPS-denied Environment\\
}
\author{\IEEEauthorblockN{Ganesh Sapkota}
\IEEEauthorblockA{\textit{Department of Computer Science} \\
\textit{Missouri University of Science and Technology}\\
Rolla, MO, USA \\
gs37r@mst.edu}
\and
\IEEEauthorblockN{Sanjay Madria}
\IEEEauthorblockA{\textit{Department of Computer Science} \\
\textit{Missouri University of Science and Technology}\\
Rolla, MO, USA\\
madrias@mst.edu}
}

\maketitle

\begin{abstract}
In modern battlefield scenarios, the reliance on GPS for navigation can be a critical vulnerability. Adversaries often employ tactics to deny or deceive GPS signals, necessitating alternative methods for the localization and navigation of mobile troops. Range-free localization methods such as DV-HOP rely on radio-based anchors and their average hop distance which suffers from accuracy and stability in dynamic and sparse network topology. Vision-based approaches like SLAM and Visual Odometry use sensor fusion techniques for map generation and pose estimation that are more sophisticated and computationally expensive as well. This paper proposes an efficient framework that integrates a novel landmark-based localization (LanBLoc) system with an Extended Kalman Filter (EKF) to predict the future state of moving entities along the battlefield. Our framework utilizes safe trajectory information generated by the troop control center for ground forces by utilizing identifiable landmarks and pre-defined hazard maps. The framework performs point inclusion tests on the convex hull of the trajectory segments to ensure the safety and survivability of a moving entity and determines the next point forward decisions. We present a simulated battlefield scenario for two different approaches that guide a moving entity through an obstacle and hazard-free path. We evaluated the observed trajectories given by two different approaches in terms of Average Displacement Error(ADE), Final Displacement error (FDE), and percent error of trajectory length. Using the proposed method, we observed a percent error of 6.51\% lengthwise in safe trajectory estimation with an Average Displacement Error(ADE) of 2.97m and a Final Displacement Error(FDE) of 3.27m.
The results demonstrate that our approach not only ensures the safety of the mobile units by keeping them within the secure trajectory but also enhances operational effectiveness by adapting to the evolving threat landscape. 
\end{abstract}


\section{Introduction}
In the context of military operations, the secure navigation of mobile troops in battlefield regions is a critical and complex task. Secure navigation in battlefield regions is a multifaceted concept that integrates advanced technological solutions, strategic resource management, and comprehensive security measures to address various operational challenges. It demands a precise and reliable positioning system\cite{positioning}, path planning\cite{path_planning}, and maneuver decision support systems\cite{decision_making} under potentially adversarial conditions. Positioning algorithms are central to secure navigation, especially in military operations, where precise location, timing, and coordination are critical. These algorithms determine the position of moving troops, ground vehicles, or any asset, often in challenging and dynamic environments. Several algorithms and methodologies have been developed to address object localization challenges, including GPS, Simultaneous Localization and Mapping (SLAM) \cite{ref_slam}, Visual Odometry \cite{vo_review} and other non-GPS-based techniques like DV-Hop Localization \cite{range_free_localization,dvhop_aps} in sensor networks.GPS (Global Positioning System) provides reliable location data and is widely used for military navigation\cite{gps_military}. However, its dependency on satellite signals makes it vulnerable to jamming and spoofing in combat scenarios\cite{gps_jamming,gps_drawback}.
SLAM techniques are crucial in environments where GPS is either unreliable or unavailable. They involve the creation of a map of an unknown environment while simultaneously keeping track of the individual's or vehicle's location within that environment. While SLAM (Simultaneous Localization and Mapping) and Visual Odometry are powerful techniques for navigation, especially in GPS-denied environments, they come with certain downsides and limitations, particularly in the context of secure battlefield navigation. SLAM and visual odometry algorithms, particularly those that process high-resolution data or complex 3D environments, can be computationally intensive. This requires significant processing power\cite{slam_survey}, which might be a constraint in smaller, battery-powered, or rapid-response systems. Also, they require multiple sensor fusion techniques\cite{sensor_fusion} which makes it not feasible in resource-constrained settings like battlefields. \\
By combining the landmark-based localization algorithm, and Bayesian filter-based future state prediction model, and applying geometrical constraints for controlled maneuvering, our research focuses on building a solution that helps to guide moving military troops, ground vehicles, or reconnaissance units securely along a safe trajectory in a non-GPS environment as shown in Fig.\ref{battlefield_map}. The main contributions of our research are:
\begin{figure}[!ht]
\centering
\includegraphics[width=0.45\textwidth]{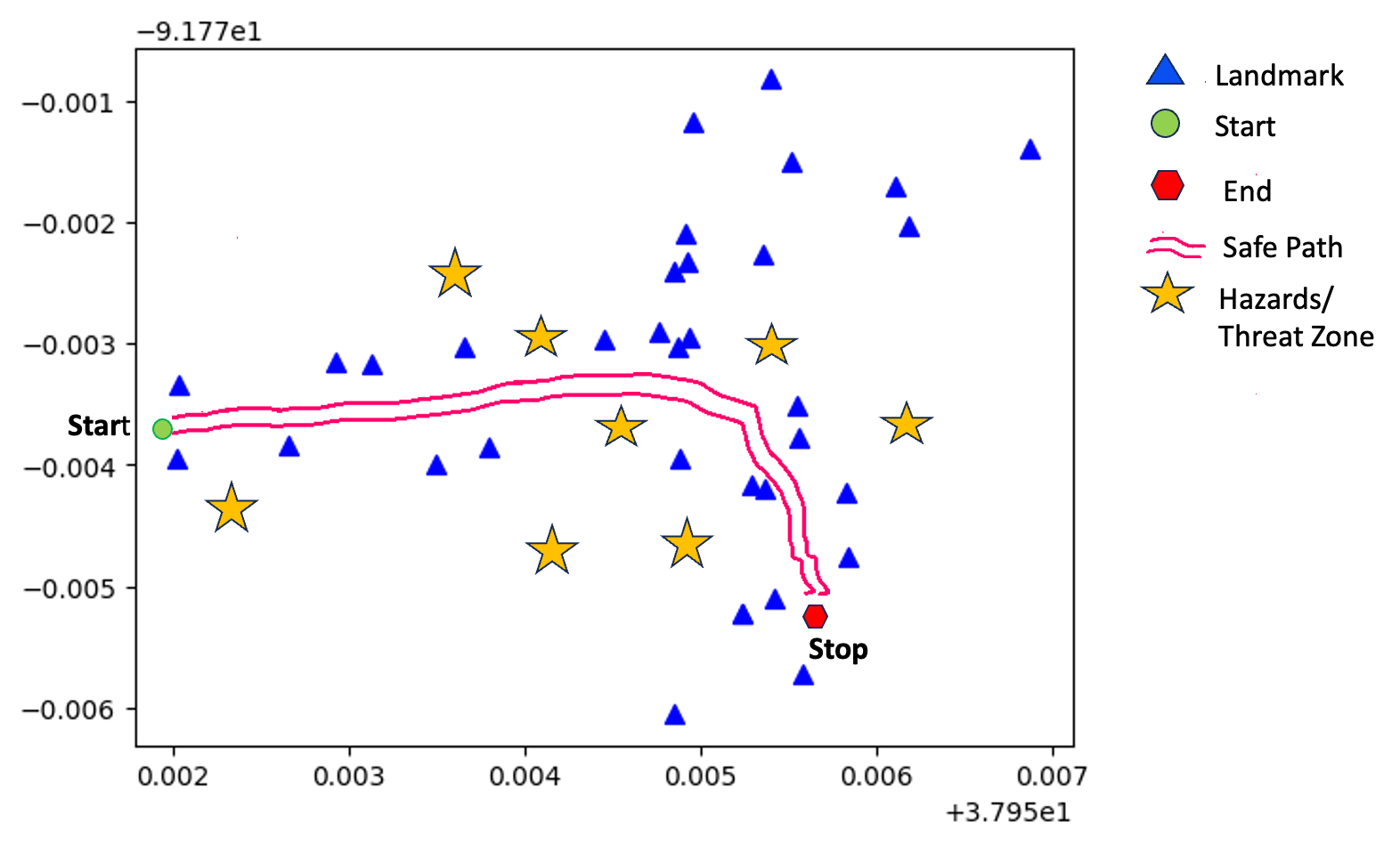}
\caption{Showing an example Safe Path in the Battlefield Map containing landmarks, obstacles, hazards components} 
\label{battlefield_map}
\end{figure}
\begin{enumerate}
    \item We proposed a non-linear motion model of a mobile entity (unmanned ground vehicle, military force or reconnaissance squadron, etc.) unit on the battlefield considering the uncertainty and variability of motion kinematics due to factors like terrain and other components such as obstacles and hazards
    \item We combined the battlefield motion model and landmark based location measurement model with an Extended Kalman filter and modeled an adapted Extended Kalman Filter to precisely estimate the state of the moving entity.
    \item We proposed the geometric method-based algorithm to safely guide the mobile entities along a predefined safe trajectory that avoids hazards and obstacles while maneuvering toward the secure destination.
    \item We observed a percent error of 6.51\% lengthwise in safe trajectory estimation with an Average Displacement Error(ADE) of 2.97m and a Final Displacement Error(FDE) of 3.27 m using the proposed Navigation approach.
\end{enumerate}
This paper is organized into five Sections. Section I introduces the research problem, and motivation then highlights our contributions to solving the problem. Section II discusses Related Works. Section III describes our approach and explains the system overview in detail. Section IV discusses the Experimental setup, Dataset, performance results, and comparison with existing works. Section V concludes the paper with the application and future direction of research.\\
\section{Background and Related Works}
Recent advancements in GPS-free positioning, including Simultaneous Localization and Mapping $(SLAM)$ and visual Odometry, have significantly improved the navigation and mapping capabilities of autonomous systems and unmanned ground vehicles, particularly in challenging indoor or GPS-denied environments.

DV-hop-based range-free localization algorithms commonly used in wireless sensor networks $(WSNs)$\cite{dvhop_aps,dv_hop_Cyclotomic,dvhop_manhattanD,csdvhop,dvhop_nsgaii,dvhop_ss} have received enormous attention due to their low-cost hardware requirements for nodes and simplicity in implementation. 
The algorithm estimates the location of unknown nodes based on their hop count to known anchor nodes and the known location of the anchor nodes. However, these methods only perform well in static, isotropic, and uniformly distributed network topology. They are sensitive to node mobility and cannot adapt to dynamic network conditions.

Vision-based methods\cite{ref_slam,ref_orbslam,reforbslam2,orb_slam_object_detection}, for robot localization, have achieved much attention in recent years due to their ability to provide precise and reliable localization results in GPS-denied environments. They utilize image and video data to determine the robot's location within its surroundings. Simultaneous Localization and Mapping (SLAM) technique ~\cite{ref_slam} estimates the position and orientation $(pose)$ of a robot within an environment while simultaneously creating its map. However, this technique requires a fusion of multiple cameras, lidar, or radar sensors to generate a map of its surroundings. Visual Odometry $(VO)$ methods \cite{deepLocalization,kittiVO} also estimate the robot's motion by tracking the changes in images or video frames captured by the robot. Structure-from-Motion algorithm (SfM)\cite{ref_sfm} reconstructs the 3D structure of the scene from multiple images which is useful for outdoor environments with well-defined features. Deep Learning-based Localization\cite{ref_pose_estimate_yolo}, particularly convolutional neural networks (CNNs), have demonstrated impressive performance in vision-based localization. CNNs can extract high-level features from images and videos, enabling robust localization in various environments.\\
For linear systems with Gaussian noise, the Kalman Filter\cite{kalman_filter} provides a framework for predicting future states of moving objects in addition to estimating the current state and is computationally efficient for real-time applications.  However, KF is only optimal for linear systems whose performance degrades with non-linearity. The nonlinear systems are better handled by other Bayesian filters such as Extended Kalman filter $(EKF)$, Unscented Kalman filter$(UKF)$, and Particle filter$(PF)$. EKF extends KF to handle nonlinear systems by linearizing them at each time step by calculating Jacobians of estimating functions. It is more popular in many applications, especially in robotics and aerospace, for tasks like localization and tracking\cite{ekf0,ekf_1,ekf_2}. PFs are also suitable for non-Gaussian and highly nonlinear systems\cite{pf1,pf2}. However, they can be computationally expensive if the right number of particles is not selected. So, It requires careful tuning of the right number of particles and the resampling strategy which can eventually affect the filter's performance. UKF also handles linearization by using a deterministic sampling approach, providing a more accurate approximation for nonlinear systems. UKF can handle arbitrary noise distributions more effectively than EKF. However, UKF is more computationally intensive than EKF due to the sigma point calculation and propagation. Selecting appropriate parameters for the sigma points can be challenging and may require empirical tuning.
Among the other filters, the Extended Kalman Filter $(EKF)$ is particularly feasible and efficient for use in battlefield scenarios due to its ability to handle non-linear systems, which are commonly encountered in military applications. For typical battlefield scenarios, EKF strikes a good balance between accuracy, computational efficiency, and robustness, making it a widely used choice in military applications.\\
The convex hull technique can be effectively used to find the boundary of a given safe trajectory, particularly in contexts like navigation, robotics, or autonomous vehicles\cite{convex_hull_app1}. The convex hull of a set \( S \) consisting of \( N \) points in an \( m \)-dimensional space is defined as the intersection of all convex sets that contain \( S \). The convex hull \( C \) of points \( p_1, ..., p_N \) is defined by \cite{convex_hull_app2} as below: 
\begin{equation}
    C = \left\{ \sum_{j=1}^{N} \lambda_j p_j \mid \lambda_j \geq 0, \sum_{j=1}^{N} \lambda_j = 1 \right\} 
\end{equation}\\
where $\lambda$ represents a set of coefficients in the linear combination of points that form the convex hull.
\begin{algorithm}
\caption{Chan's Algorithm for Convex Hull}
\label{algo_chan}
\begin{algorithmic}[1]

\Require Set of points $P$
\Ensure Convex hull of $P$
\State Initialize $m \leftarrow \text{initial guess for } h$, the number of
vertices of convex hull polygon
\State Divide $P$ into subsets $P_1, P_2, \ldots, P_{\lceil \frac{n}{m} \rceil}$ each of size at most $m$
\ForAll{subset $P_i$}
    \State Compute convex hull $H_i$ using \textbf{Algorithm \ref{algo_ghram_scan}}
\EndFor

\While{}{true}
    \State Initialize $H \leftarrow$ convex hull of leftmost point
    \ForAll{$i \leftarrow 1$ to $m$}
        \ForAll{subset convex hull $H_j$}
            \State Find point $p \in H_j$ that is a tangent to $H$ through last added point
            \State Add $p$ to $H$
        \EndFor
        \If{all points in $H$ are found and hull is closed}
            \State $H$ as the convex hull of $P$
        \EndIf
    \EndFor
    \State Double the value of $m$ and repeat
\EndWhile
\end{algorithmic}
\end{algorithm}
In a 2-dimensional space, a convex hull is the smallest convex set that completely encloses a given set of points \( S \). In robotics, convex hulls are used in motion planning and obstacle avoidance algorithms to simplify and solve spatial problems. Convex hulls can be used in GIS applications to analyze geographical data, like defining the boundary of a set of geographical locations. 
Jarvis march algorithm \cite{jarvis_algo}, computes the convex hull with a time complexity of $O(nh)$, where $n$ and $h$ represent the number of points and nodes of the hull polygon. Graham's Scan \cite{ghrams_scan} shown in \ref{algo_ghram_scan}is another efficient method for computing the convex hull in 2D with a time complexity of $O(n\log n)$. The algorithm selects the point with the lowest y-coordinate and then sorts the remaining points based on the angle they make with the initial point and the x-axis. It finally obtains the hull by traversing the sorted array and ensuring that each new point forms a left turn. 
Chan’s algorithm \cite{chans_algo} is shown in \textbf{Algorithm \ref{algo_chan}} with the time complexity of $O(n \log h)$, where $n$ indicates the total number of points in a cluster and $h$ indicates the number of vertices of convex hull polygon. The algorithm divides the set of given points into smaller subsets, finds the convex hull on each subset, and then merges those hulls. The merging phase utilizes the idea of the tangent finding process similar to the Graham scan algorithm shown in \textbf{Algorithm \ref{algo_ghram_scan}}.

\begin{algorithm}
\caption{Graham's Scan Algorithm for Convex Hull}
\label{algo_ghram_scan}
\begin{algorithmic}[1]
\Require Set of points $P$
\Ensure Convex hull of $P$
\State Find the point $p_0$ with the lowest y-coordinate or the leftmost in case of tie
\State Sort the points in $P$ by polar angle with $p_0$
\State Initialize stack $S$ and push $p_0$, $p_1$, and $p_2$ onto $S$
\ForAll{$i \leftarrow 3$ to $n-1$}
    \While{the angle formed by points next-to-top($S$), top($S$), and $p_i$ makes a non-left turn}
        \State pop $S$
    \EndWhile
    \State push $p_i$ onto $S$
\EndFor
\State $S$ as the convex hull of $P$
\end{algorithmic}
\end{algorithm}
The convex hull technique provides a geometrically sound and computationally efficient method to determine the boundaries of a safe trajectory, making it a valuable tool in navigation and path-planning applications. Our method applies the convex hull approach to check the safe zone while making the maneuvering decision by the moving entity while navigating.

\section{Our Methodology and System Model}
\subsection {System Overview}
\begin{figure}[!ht]
\centering
\includegraphics[width=0.35\textwidth]{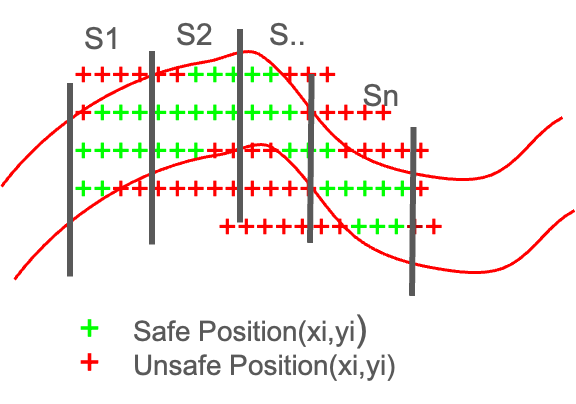}
\caption{Safe trajectory as a series of path segments} 
\label{safe_positions}
\end{figure}

Our method is designed to guide a mobile troop through an obstacle and hazard-free path starting from its current position to the destination. Moving troops run Localization algorithm $(LanBLoc)$ to obtain its current position and communicate with the control center using a secure communication protocol. The node obtains the safe trajectory as a series of path segments $S = (s_1,s_2...s_n)$ as shown in Fig. \ref{safe_positions}, each containing trajectory points $p_i = (x_{i..n},y_{i..n})$ from the control center via secure communication. For each path segment $s_i = (p_1,p_2,...p_n)$, the node computes the convex hull $h_i$ of the points cluster using Algorithm \ref{algo_chan} and stores them in sorted order as $H = (h_1,h_2....h_n)$.

\begin{figure}[!ht]
\centering
\includegraphics[width=0.5\textwidth]{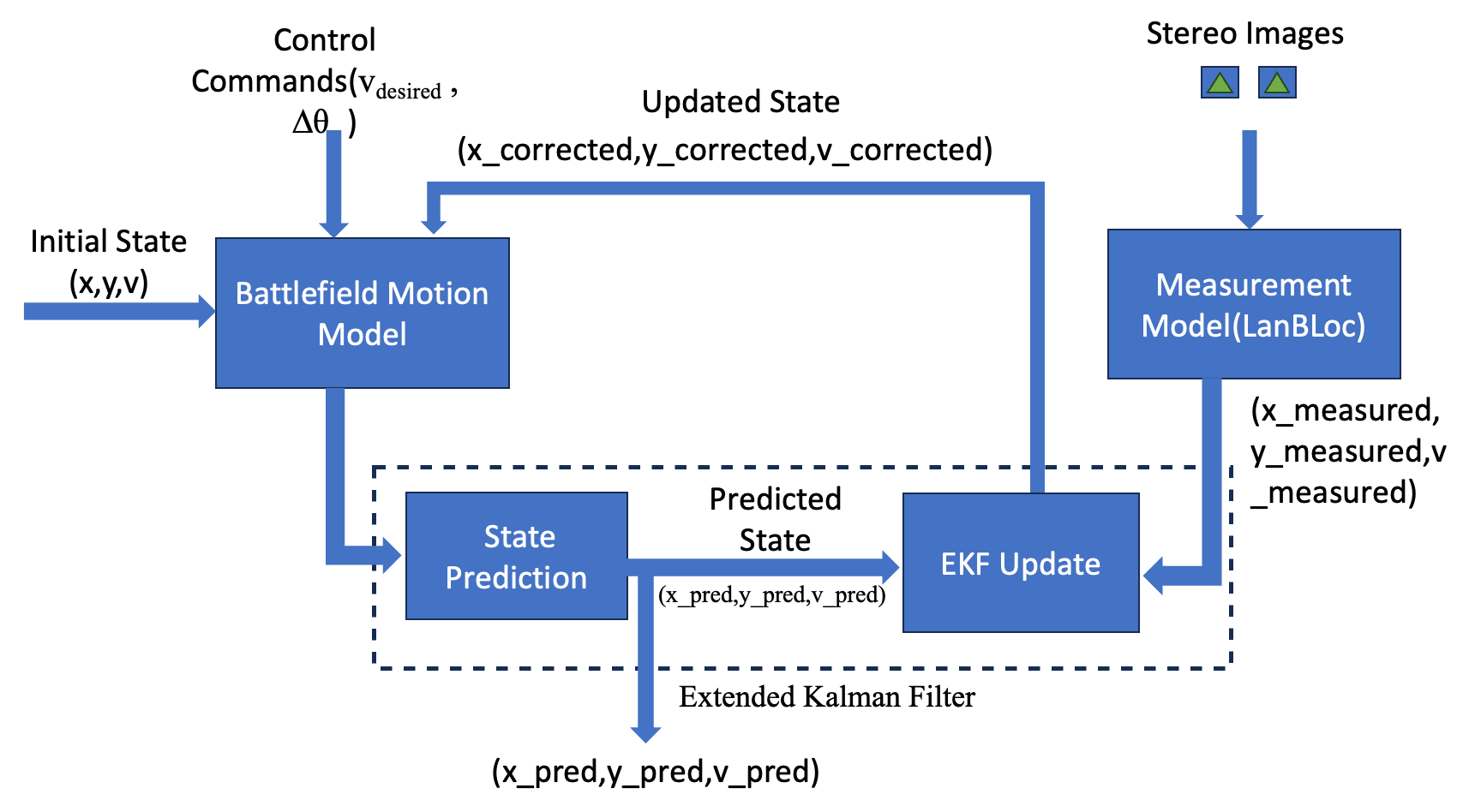}
\caption{Extended Kalman Filter adapted based on battlefield motion model} 
\label{syst_overview}
\end{figure}
The command and control server generates the safe path shown in Fig. \ref{safe_positions} and provides information to the moving entity. Moving entities have to navigate along this safe path to ensure the safety, survivability, and success of the mission. Moving entity follows a motion model and control commands for heading decisions and maneuvering. Fig. \ref{syst_overview} shows the precise state prediction model of a moving entity on the battlefield using the Extended Kalman filter provided with the LanBLoc algorithm as a measurement model Shown in Fig.\ref{lanbloc_architecture} discussed in detail in Section III, Subsection B.\\
At the beginning of each trial, the node obtains its current position using the LanBLoc algorithm \ref{algo_StereoLocalization} and velocity from the measurement sensor which serves as an input for the Extended Kalman Filter that predicts the next state $(x_{pred},y_{pred},v_{pred})$ based on the motion model. Then the node checks if the predicted position lies within the convex hull which indicates the safe zone and makes the heading decisions. If the position lies within the convex hull, it will move with the default command (desired speed, desired heading/direction). Otherwise, it checks the adjacent convex hulls and makes movement decisions. The EKF-based trajectory prediction mechanism is explained in detail in Section II, Subsection D.
\subsection{Landmark-based moving object localization (LanBLoc)}
\begin{figure}[!ht]
\includegraphics[width=0.5\textwidth]{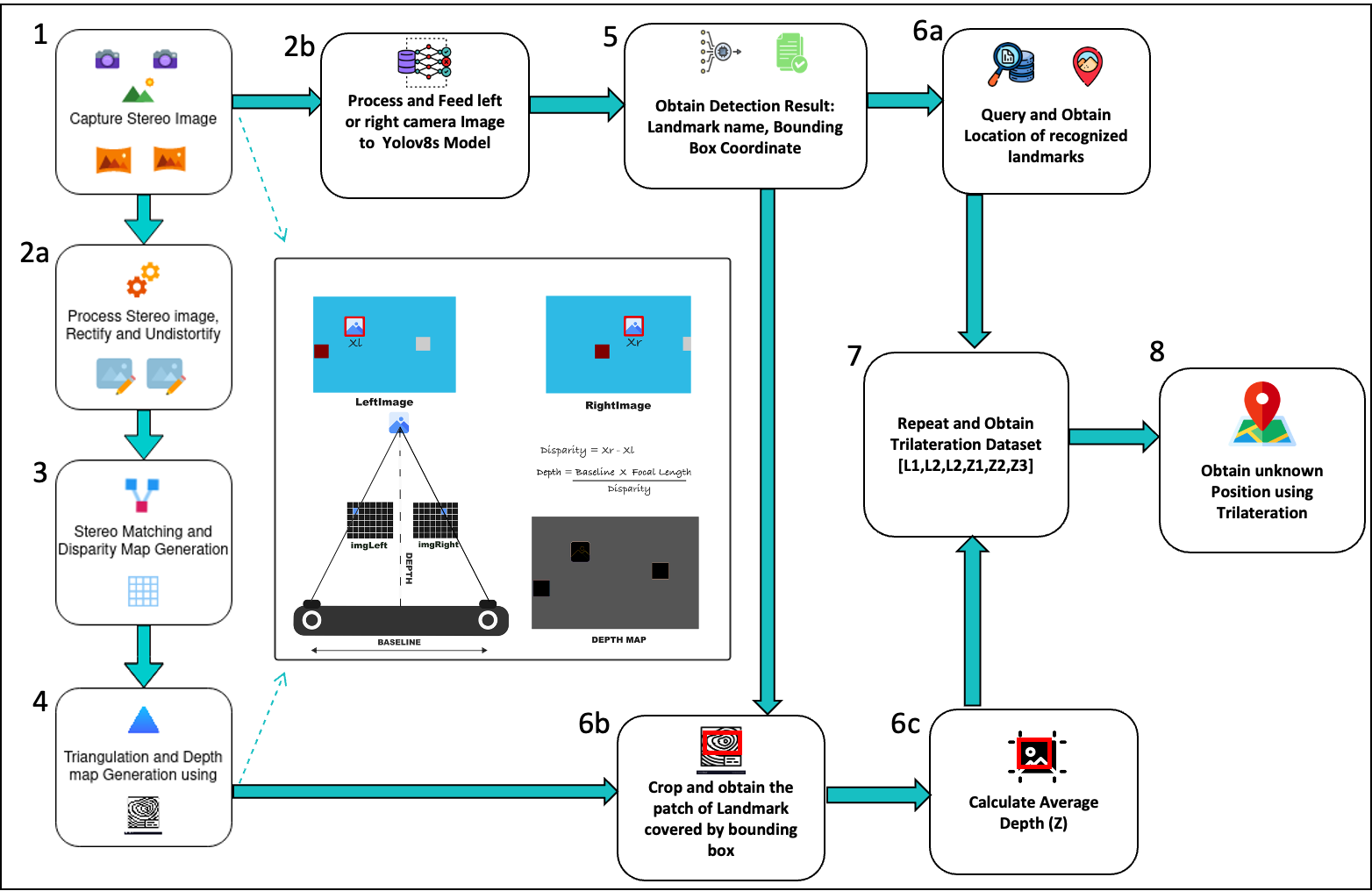}
\centering
\caption{System Overview of LanBloc framework. Distance estimation and landmark recognition steps and the results after two operations are fused to calculate the position of an unknown node.} 
\label{lanbloc_architecture}
\end{figure}
Deploying anchor nodes in Wireless Sensor Networks (WSNs) can be particularly demanding, especially in harsh environments like battlefields, where network connectivity is often sparse and unstable. In such scenarios, traditional anchor nodes struggle to maintain communication and update critical localization information due to their limited coverage and the dynamic nature of the network. Additionally, the energy and bandwidth costs associated with broadcasting beacon packets from physical anchors are high. To address these challenges, our research presents a novel solution shown in Algorithm \ref{algo_StereoLocalization}  by utilizing the existing physical and geographical landmarks as anchor points.  These landmarks are first identified and precisely located using the YOLOv8-based deep learning model. A pre-trained YOLOv8s model is customized and fine-tuned using the MSTLandmarkv1 dataset to detect and recognize 34 different landmarks. Our landmark recognition model shows a Box loss precision (Pr) of $0.939$, a Recall (R) of $0.963$, and a mean Average Precision $(mAP~@~0.5~IoU)$ score of $0.957$ and $mAP~@~[0.5:0.95] ~ IoU$ of $0.823$ for all landmark classes.\\
The distances between the moving entity and the identified landmark anchors are calculated using an efficient stereo-matching technique with high accuracy. Using the localized landmark anchors and the calculated distances to the anchors, the LanBLoc algorithm computes the position of a moving object with an average RMSE of 0.0142 m in the X-position and 0.039 m in the Y-position.  By utilizing the recognized landmarks as anchor nodes for trilateration, we enhance the accuracy and reliability of moving node localization with further optimization of position as shown in Fig.\ref{lanbloc_architecture}. Unlike traditional radio-based anchor nodes, physical landmarks are robust and enduring, making them ideal for challenging environments. Furthermore, this approach extends the applicability of our framework to sparse sensor networks deployed across extensive geographical areas. In essence, this approach offers a more practical and resilient approach to anchor node deployment and localization, particularly in rugged settings like battlefields, where deploying and maintaining physical anchors are impractical as well as dangerous. 
\begin{algorithm}
\caption{Landmark Based Localization Algorithm}
\label{algo_StereoLocalization}
\begin{algorithmic}[1]
\Require Stereo images of a scene $(i_l,i_r)$
\Ensure Current position of object $(x,y)$
\Procedure{LanBLoc}{}
    \State Capture stereo images  $(i_l,i_r)$ of a scene with a landmark Anchor
    \State Process stereo images (Normalization, White balancing, Sharpening)
    \State Rectify and undistort images using camera calibration parameters
    \State Initialize empty arrays for trilateration data: $L = [ ]$, $D = [ ]$
    
    \While{Scanning the Scene}
        \State Detect landmarks using YOLOv8s: $[(l_i, BBox_i)]$
        \ForAll{detected landmarks}
            \State Query landmark database to obtain location $L_i$
            \State Crop landmark region patch from depth map
            \State Aggregate depth levels to obtain average distance $d_i$
            \State Append $l_i$ to $L$ and $d_i$ to $D$
        \EndFor
    \EndWhile
    
    \State Perform trilateration to obtain localization:
    \State $[l_1, l_2, l_3, d_1, d_2, d_3] = [L[0], L[1], L[2], D[0], D[1], D[2]]$
    
    \State Use the least square method to estimate the position (x,y)
    \State Optimize the position using L-BFGS-B optimization method
    \State \textbf{Output:} Optimized position (x,y)
\EndProcedure
\end{algorithmic}
\end{algorithm}

\subsection{Motion Model of Mobile Entity in the Battlefield}
Creating a motion model for a battlefield scenario involves considering both the physical aspects of motion and the tactical constraints imposed by the battlefield. Our model considers the vehicle's or troop unit's velocity, acceleration, and deceleration limits, and its maneuverability, which is how quickly it can change its direction. Given a control command and current state, a predicting function updates the unit's next state based on those factors.  The motion model of a moving entity on a battlefield can be represented in mathematical form as shown in Eqn: \ref{vel_update},\ref{head_update},\ref{pos_update}. This model incorporates kinematic aspects of a moving object such as position, velocity, and heading, as well as dynamic control inputs(commands) for desired speed and heading change. Let's denote:\\
\(\vec{p}_t = [x_t, y_t] \) as the position vector at time \( t \).\\
\( v_t \) as the velocity at time \( t \).\\
\( \theta_t \) as the heading (direction) at time \( t \).\\
\( v_{\text{desired}} \) as the desired velocity (speed command).\\
\( \Delta\theta \) as the change in heading (heading command).\\
\( a \) as the acceleration capability of the unit.\\
\( d \) as the deceleration capability of the unit. \\
\( m \) as the maneuverability (maximum rate of change of heading).\\
\( \Delta t \) as the time step.\\
Now, the motion model can then be expressed as:
\subsubsection{Velocity Update:}
\begin{equation}
\label{vel_update}
 v_{t+1} = 
   \begin{cases} 
   \min(v_t + a\Delta t, v_{\text{desired}}, v_{\text{max}}) & \text{if } v_{\text{desired}} > v_t \\
   \max(v_t - d\Delta t, v_{\text{desired}}, 0) & \text{if } v_{\text{desired}} \leq v_t 
   \end{cases}
\end{equation}
where, \( v_{\text{max}} \) is the maximum speed of the unit.

\subsubsection{Heading Update:}
\begin{equation}
\label{head_update}
\theta_{t+1} = \theta_t + \text{clip}(\Delta\theta, -m\Delta t, m\Delta t)
\end{equation}
    The function \( \text{clip}(x, a, b) \) restricts \( x \) to the range \( [a, b] \).
    
\subsubsection{Position Update:}
\begin{equation}
\label{pos_update}
 \vec{p}_{t+1} = \vec{p}_t + v_{t+1}\cdot \Delta t \cdot \begin{bmatrix} \cos(\theta_{t+1}) \\ \sin(\theta_{t+1}) \end{bmatrix}
\end{equation}

In this model, the velocity update ensures that the velocity at the next time step, \( v_{t+1} \), is updated based on the current velocity, \( v_t \) while respecting the acceleration/deceleration capabilities and the maximum/minimum speed limits. The heading update adjusts the heading, \( \theta_{t+1} \), by a controlled amount, \( \Delta\theta \), within the limits imposed by the unit's maneuverability. The position update calculates the new position, \( \vec{p}_{t+1} \), based on the updated velocity and heading. The displacement is broken down into x and y components using trigonometric functions, considering the unit's heading. This model assumes a linear motion and instant change of speed and heading for simplicity.\\
However, in practice, the dynamics of the battlefield are unpredictable due to changing environmental conditions, terrain, and obstacle components. To represent the motion model of the real-battlefield scenario, we need to consider the process noise in position, heading, and velocity, that capture the uncertainty and variability due to factors like terrain and other unmodeled influences such as obstacles and hazards. It includes terms for terrain effects $(\tau)$ and process noises ${\epsilon}_{x,t}, {\epsilon}_{y,t},{\epsilon}_{\theta,t},{\epsilon}_{v,t}$ in position, heading, and velocity respectively. Hence motion model is defined in Eqns. \ref{vel_update},\ref{head_update},\ref{pos_update} is updated as:
\begin{equation}
\label{vel_update2}
    v_{t+1} = f(v_t, a_t, \tau, \epsilon_{v,t}) 
\end{equation}

\begin{equation}
\label{head_update2}
     \theta_{t+1} = \theta_t + \omega_t \cdot \Delta t + \epsilon_{\theta,t} 
\end{equation}

\begin{equation}\label{pos_update2}
    \begin{split}
        x_{t+1} = x_t + v_{t+1} \cdot \Delta t \cdot \cos(\theta_{t+1})  + \epsilon_{x,t}
        \\y_{t+1} = y_t + v_t \cdot \Delta t \cdot \sin(\theta_{t+1}) + \epsilon_{y,t}
        \\\vec{p}_{t+1} = \vec{p}_t + v_{t+1}\cdot \Delta t \cdot \begin{bmatrix} \cos(\theta_{t+1}) \\ \sin(\theta_{t+1}) \end{bmatrix} + \begin{bmatrix} \epsilon_{x,t}\\ \epsilon_{y,t} \end{bmatrix}
    \end{split}
\end{equation}
where ${\omega}_t$ is the angular velocity (rate of heading change), and $f$ is a function that computes the next velocity based on the current velocity, acceleration, and terrain. 
The above representation captures the essential dynamics of the motion model for any moving entities for eg: a ground vehicle or troop unit on a battlefield, taking into account both their physical capabilities and the control inputs for navigation.\\
To guide the object from a start position to the end position given a pre-defined safe path, the following steps are performed utilizing the above motion model.
\begin{enumerate}
    \item Set initial position $(x_i,x_j)$ 
    \item Get Initial Control Input $U_i\leftarrow(v_i,{\theta}_i)$ using following steps.
    \begin{enumerate}
        \item Consider arbitrary current velocity ($v_i$).
        \item Compute the time $t$ to reach the centroid $(x_j,y_j)$ of subsequent path segment based on distance and current velocity: $t_c = d_c / v_i$, where $d_c$ is calculated as, $d_c = \sqrt{(x_i-x_j)^2 + (y_i-y_j)^2}$.
    \end{enumerate}
    \item Calculate next possible position after $t$ seconds based on the control input $U$.
    \item Check if the predicted position is within the current path segment.
    \begin{enumerate}
        \item If the position is within the segment:
        \begin{enumerate}
            \item Move the object to that position with the control input calculated at current time.
            \item Append the predicted position to the trajectory list.
            \item When the object travels to the new position, calculate the control input based on the previous and current position, then repeat the process until the object reaches near the centroid (some threshold value) of the last cluster.
        \end{enumerate}
        \item Otherwise:
        \begin{enumerate}
            \item Stop moving to the predicted position.
            \item Get control input to move to the centroid of the next cluster/segment for fineness using the method in Step 2.
            \item Repeat Steps 3 and 4.
        \end{enumerate}
    \end{enumerate}
\end{enumerate}

\subsection{Trajectory Prediction using Extended Kalman Filter based battlefield motion model}
On the battlefield, position tracking and navigation systems are often non-linear due to complex terrain, dynamic obstacles, and hazardous components that should be avoided by moving entities to ensure their survivability.
The Extended Kalman Filter (EKF) is an extension of the Kalman filter for nonlinear systems. It linearizes the current mean and covariance, allowing it to approximate the state of a nonlinear system. For the non-linear battlefield scenario, an Extended Kalman Filter (EKF) can be an efficient and effective filtering algorithm due to its low computational complexity and high accuracy of estimation. 
Our main approach is to adapt the Extended Kalman Filter by integrating our battlefield motion model defined in Eqns.\ref{vel_update2},\ref{head_update2}, and \ref{pos_update2} in its prediction step to predict the future state of a moving entity. The mathematical representation of EBKF is shown below:
\subsubsection{State and Measurement Models}
The state of the system at time step $k$ is represented by the vector $x_k$. The EBKF models the system dynamics and measurement with the following nonlinear functions:
The state transition function $X_{k+1}$ and the measurement function $Z_{k+1}$ are represented as in Eqn. \ref{state_transition_func} below:
    \begin{equation}
    \label{state_transition_func}
    \begin{split}
        X_{k+1} = f(X_{k},U_{k})+ W_{k}\\
        Z_{k+1} = h(X_{k+1})+V_{k+1}
    \end{split}
    \end{equation}
$f(\cdot)$ is the nonlinear state transition function, which predicts the next state based on control input $U_k$ and the previous state $X_k$ while the $h(\cdot)$ is the measurement function, that maps the true state space into the observed space.$W_k$ represents the process noise and $V_{k+1}$ refers to the measurement noise which are assumed to be normally distributed with covariance $Q_k$ and $R_{k+1}$ respectively. The battlefield navigation adopts the \textbf{LanBLoc} shown in Algorithm \ref{algo_StereoLocalization} as a measurement function.

\subsubsection{Prediction Step}
The EKF predicts the state and error covariance at the time interval $k$. The predicted state estimate $\hat{X}_{k+1}$ and  and predicted covariance estimate $P_{k+1}$ are given by: 
\begin{equation}
\label{predicted_state}
\begin{split}
      \hat{x}_{k+1}=f(\hat{x}_k,U_k) + w_{k}\\
     P_{k+1}= F_k \cdot P_k \cdot F_{k}^T+ Q_k
\end{split}
\end{equation}
where $F_k$ is the Jacobian of state transition function $f(\cdot)$ with respect to $X$ that is evaluated at $\hat{X}_{k+1}$ and $Q_k$ is the covariance matrix of process noise. The Jacobian is the matrix of all partial derivatives of the vector of $f(\cdot)$ and $h(\cdot)$ around the estimated state.
around the estimated state

\subsubsection{Update Step}
The EBKF incorporates the new location measurement calculated by algorithm \ref{algo_StereoLocalization} to update the state estimate and covariance. The measurement residual $\Tilde{y}_{k+1}$ and residual covariance $S_{k+1}$ are calculated using Eqn. \ref{residual_cov} below.\\
\begin{equation}
\label{residual_cov}
\begin{split}
     \Tilde{y}_{k+1} = z_k-h(\hat{x}_{k+1})\\
    S_{k+1}=H_{k+1}\cdot P_{k+1}\cdot H_{k+1}^T+R_{k+1}
\end{split}
\end{equation}
Where $z_{k+1}$ represents the actual measurement of the position at time step $k+1$.
The Kalman Gain is a fundamental component of the Kalman Filter that determines the degree to which the new measurement will be incorporated into the state estimate. The calculation of the Kalman Gain balances the uncertainty in the current state estimate $($as represented by the state covariance matrix$)$ and the uncertainty in the new measurement as represented by the measurement noise covariance.
So, the Kalman Gain $ K_{k+1}$ is computed using the predicted covariance estimate $P_{k+1}$, the transpose of the observation model $H_{k+1}^T$, and the inverse of the residual covariance $S_{k+1}^{-1}$ in equation \ref{kalman_gain} below.
\begin{equation}
\label{kalman_gain}
    K_{k+1}=P_{k+1} \cdot H_{k+1}^T \cdot S_{k+1}^{-1}
\end{equation}
Finally, the state estimate $\hat{x}_{k+1}$ and covariance estimate $P_{k+1}$ are updated in Eqn. \ref{state_update}.
\begin{equation}
    \label{state_update}
    \begin{split}
     \hat{x}_{k+1} = \hat{x}_k +K_{k+1}\cdot \Tilde{y}_{k+1}\\
        P_{k+1} = (I-K_{k+1}\cdot H_{k+1}) \cdot P_{k+1}
\end{split}
\end{equation}
Where $I$ is the identity matrix.

\subsection{Safety Check using Convex Hull and Point in Polygon Method}
In the context of battlefield navigation, a safe path refers to a field that avoids obstacles and possible hazards. A continuous function that maps a parameter $s$ representing a sequence to a point in the navigable space, such that the path avoids collisions with obstacles and hazards, and respects the constraints of the navigable space. Mathematically, a safe path \( \mathbf{p}(s) \) can be defined as follows:
\begin{equation}
     \mathbf{p}(s) : [s_0, s_f] \rightarrow \mathbb{R}^n
\end{equation}
where $s$ is the parameter, representing sequence, with  $s_0$ and $s_f$ being the start and end sequence numbers, $\mathbb{R}^n$ represents the n-dimensional navigable space, $ \mathbf{p}(s)$ is the position in the navigable space.
The path starts at the initial position $p_0$ and ends at the target position $pf$, which is given by:\\
\begin{equation}
    \begin{split}
        p(s_0)=p_0\\
        p(s_f)=p_f
    \end{split}
\end{equation}
The path is generated by the control center based on the current information about the state of the obstacle and hazard components within the area of interest.
To perform the safety check, the path is split into multiple segments and stored preserving their sequence information. For each segment, $S =$\{($x_1$,$y_1$),($x_2$,$y_2$),...,($x_n$,$y_n$)\}, convex hull is calculated. The predicted next state from the EKF is checked against the convex hull by adapting the even-odd test method in \cite{even_odd_algo}.
\begin{algorithm}
\caption{Safety Check based on EKF aided object movement }
\begin{algorithmic}[1]
    \Require $EKF\_predicted\_position$, $path\_segments$
    \Ensure Boolean value of point inclusion test, control input $U$
    \ForAll{$segment$ in $path\_segments$}
        \State $current\_hull \leftarrow ConvexHull(segment)$
        \State $next\_segment\_index \leftarrow$ index of next segment
        \If{$next\_segment\_index$ exists}
            \State $next\_hull \leftarrow ConvexHull(next\_segment)$
        \Else
            \State $next\_hull \leftarrow$ None
        \EndIf
        \If{PointInConvexHull($EKF\_predicted\_position$, $current\_hull$) \\
            \hspace{1.5cm} or ($next\_hull$ is not None and \\
            \hspace{1.5cm} PointInConvexHull($EKF\_predicted\_position$, $next\_hull$))}
            \State MoveToObjectPosition($EKF\_predicted\_position$)
            \State EKFUpdate()
        \Else
            \State ComputeNewControlInput($current\_hull$, $next\_hull$)
            \State break
        \EndIf
    \EndFor
\end{algorithmic}
\end{algorithm}

\section{Experiments and Results}
To test the validity of our framework, a localization experiment was performed on Alienware Aurora R12 System with 11th Gen Intel Core i7 CPU, 32 GiB Memory, and NVIDIA GeForce RTX 3070 GPU using Python 3.10 on PyCharm 2022.1 (Edu) IDE. The landmark recognition model was trained on the Google Colab environment with Python-3.10.12 and torch-2.1.0+cu118, utilizing an NVIDIA Tesla T4 GPU with 16GiB memory.
We used the following datasets for our experiments.
\subsection{Datasets}
\subsubsection{MSTlandmarkv1 Dataset}
MSTlandmarkv1 dataset was used to train the yolov8-based landmark recognition model comprising around ${4000}$ images of $34$ real-world landmark instances which were labeled manually. The images were captured using a camera with $2K$-resolution $(2560*1440px)$. The collected dataset was split into train($70$ percent), validation($20$ percent), and test($10$ percent) sets. Image augmentation was done on the training portion of the dataset to avoid overfitting problems and to improve the model's robustness. Image color jittering was done by adjusting brightness between $-25$ percent and $+25$ percent, bounding box rotation between -15° and +15°,  and bounding box noise addition up to $5$ percent of the pixels. The final dataset comprising $7547$ images including train, validation, and test sets was exported to YOLO V8 Pytorch format.
\subsubsection{MSTlandmarkStereov1 Dataset}
\begin{figure}[!ht]
\includegraphics[width=0.4\textwidth]{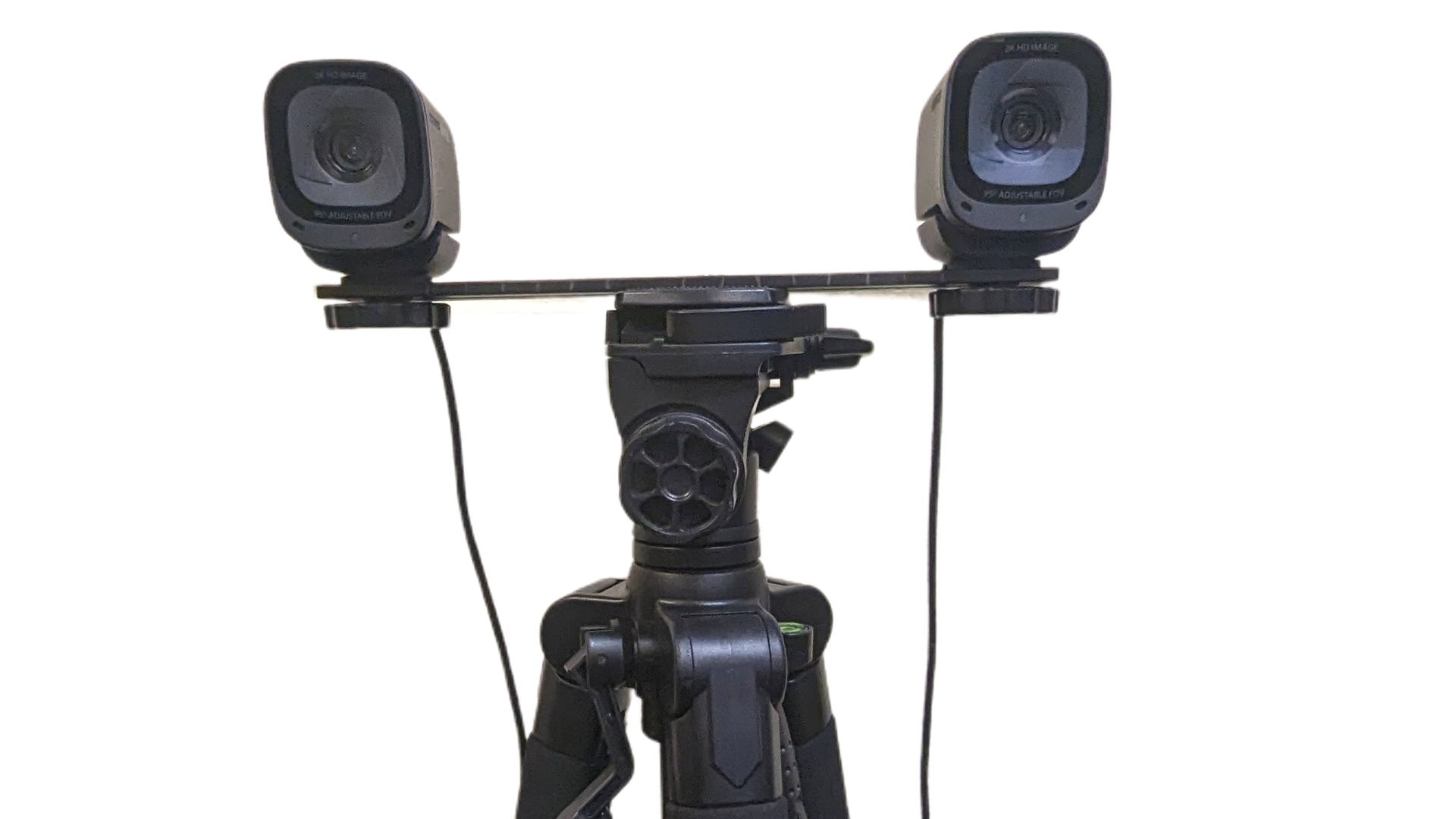}
\centering
\caption{Stereo Camera Setup for MSTlandmarkStereov1 Collection} 
\label{stereo_setup}
\end{figure}

We used a real-world landmark stereo dataset for the node-to-landmark distance estimation. We used two identical cameras of $(2560*1440px~@~30~fps)$ resolution as shown in Fig. \ref{stereo_setup} with adjustable field of view (FoV) to collect this dataset. The cameras were arranged on the metal bar with adjustable baseline(B) from $10-40$ centimeters. This dataset is used to perform distance estimation from unknown nodes to the landmark anchors.

\subsubsection{Trajectory Dataset}
\begin{figure}[!ht]
\includegraphics[width=0.5\textwidth]{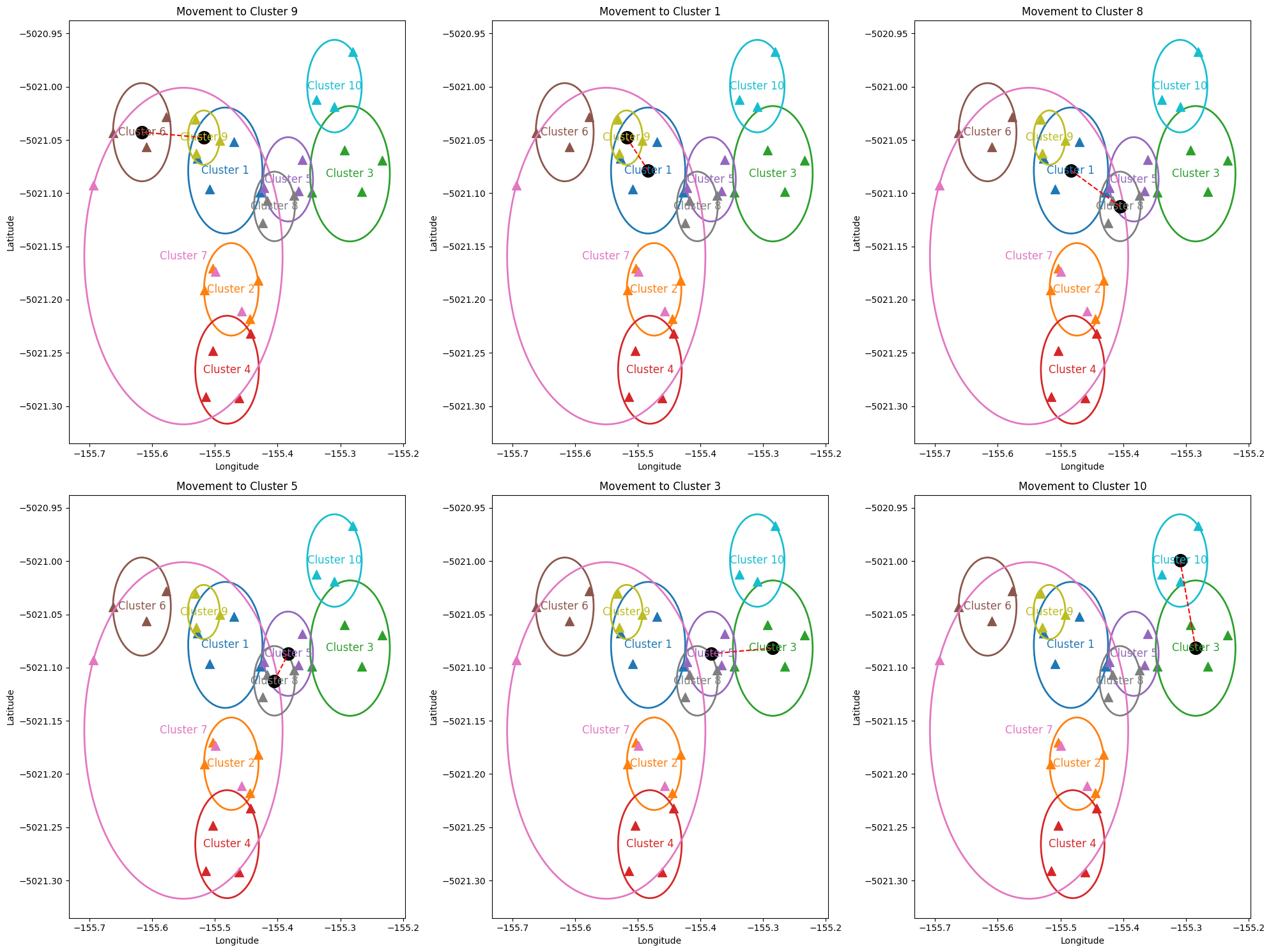}
\centering
\caption{Stepwise generation of ground truth trajectory from Cluster 6 to Cluster 10} 
\label{trace_trajectory2}
\end{figure}

\begin{figure}[!ht]
\includegraphics[width=0.5\textwidth]{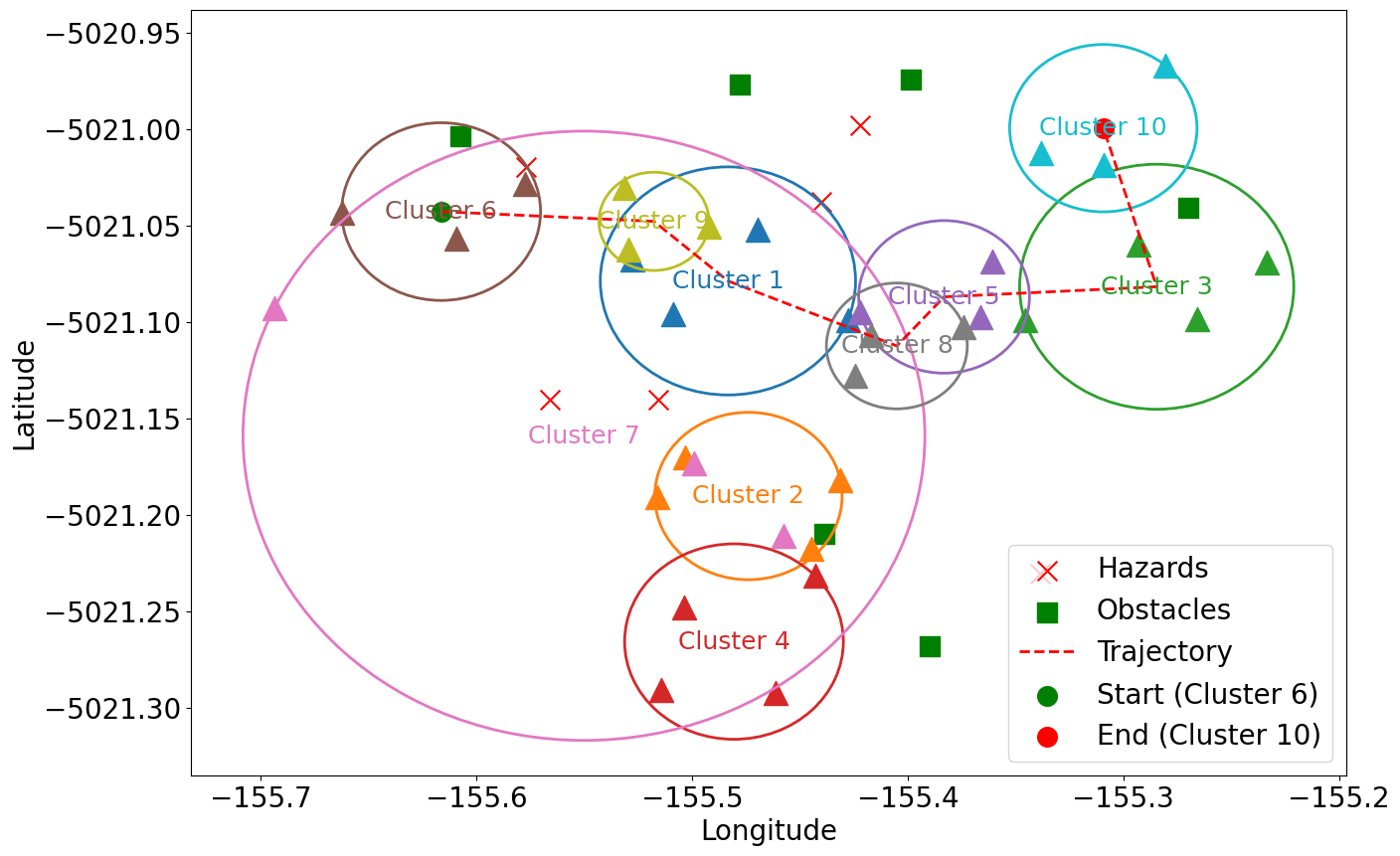}
\centering
\caption{Example showing ground truth trajectory from Cluster 6 to Cluster 10} 
\label{trace_trajectory1}
\end{figure}

We generated the safe path dataset to test our framework. We set an area of 200*200 grid space as a battlefield region and mapped all the distinct landmarks in MSTlandmarkStereov1 with their actual geo-locations. Using the information from the dataset, the landmarks are clustered and at least three landmarks that are within the detectable range of moving objects are selected from a cluster as a trilateration set to calculate their location. We first generated ground truth trajectories considering that the object passes through the centroids of different landmark clusters as shown in Fig.\ref{trace_trajectory2}. The trace of the path followed by a moving object when it travels along the centroid of each landmark cluster starting from the first cluster to the end cluster is considered the actual trajectory with the highest level of safety as shown in Fig.\ref{trace_trajectory1}. 

\begin{figure}[!ht]
\includegraphics[width=0.5\textwidth]{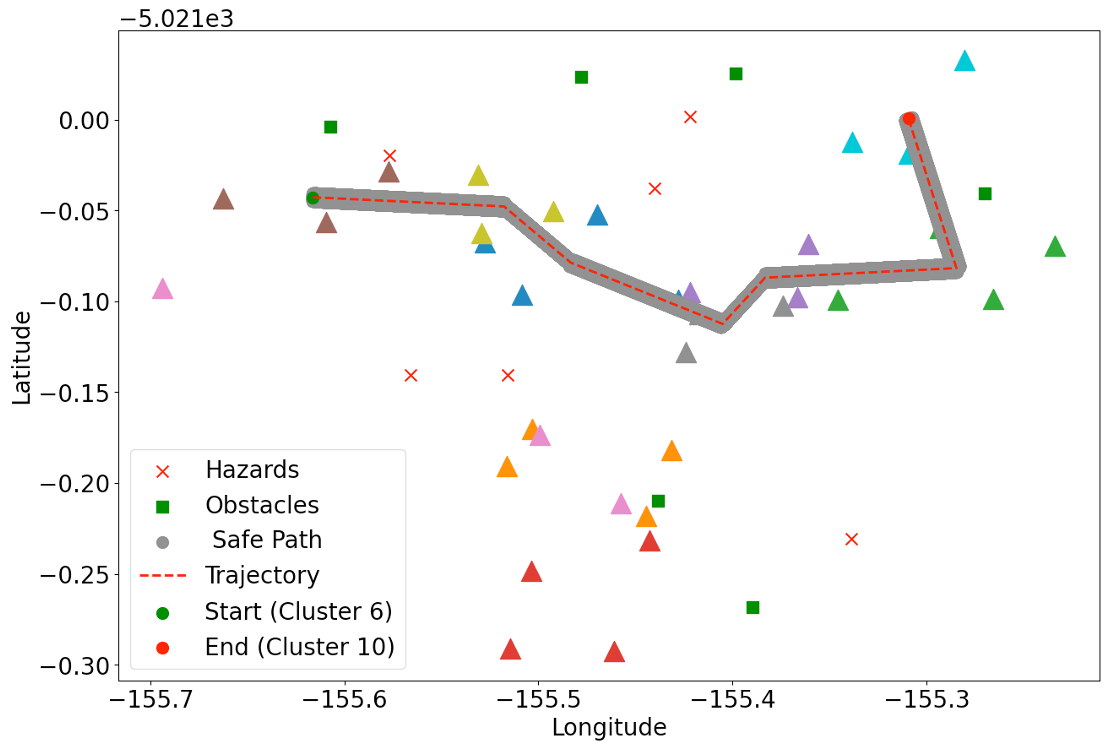}
\centering
\caption{Example Safe Path from Cluster 6 to Cluster 10.} 
\label{safe_path1}
\end{figure}

Based on actual trajectory data we generated the safe path which is obtained by defining the navigable margin on both sides of the trajectory as shown in Fig. \ref{safe_path1}. The safe path is considered as the bounded region within which moving objects can navigate safely.

\subsection{Performance Metrics}
We measured the performance of our approach using widely used performance metrics: Percent Error, Average Displacement Error(ADE), and Final Displacement Error (FDE). For each safe path, 1000 trajectories were observed and their average were calculated.\\
The percent error of the estimated trajectory indicates a discrepancy between the predicted trajectory and the actual trajectory which is calculated using Eqn. \ref{percent_error_eq}:
\begin{equation}
    \text{Percent Error} = \frac{|\text{Est. Length} - \text{True Length} |}{\text{True Length}} \times 100 \%
    \label{percent_error_eq}
\end{equation}
The ADE is an average of pointwise L2 distances between the predicted and ground truth trajectory that is calculated using Eqn. \ref{ade}.\\
\begin{equation} 
    ADE = \frac {1}{N}\sum _{t=t_{0}}^{t_{f}} \left |{\left |{{\hat {Y}_{(t)}- {Y}_{(t)}}}\right |}\right |
    \label{ade}
\end{equation} 
The final displacement error (FDE) is the L2 distance between the final points of the predicted and ground truth trajectory that is calculated using Eqn \ref{fde}.\\
\begin{equation} 
    FDE = \left |{\left |{{\hat {Y}_{\left ({t_{f}}\right)}- {Y}_{\left ({t_{f}}\right)}}}\right |}\right |
    \label{fde}
\end{equation} 
where $\hat{Y}_t$ is the predicted position at time step t and $Y_t$ is the ground truth position.

\subsection{Results}
\begin{figure}[!ht]
\includegraphics[width=0.5\textwidth]{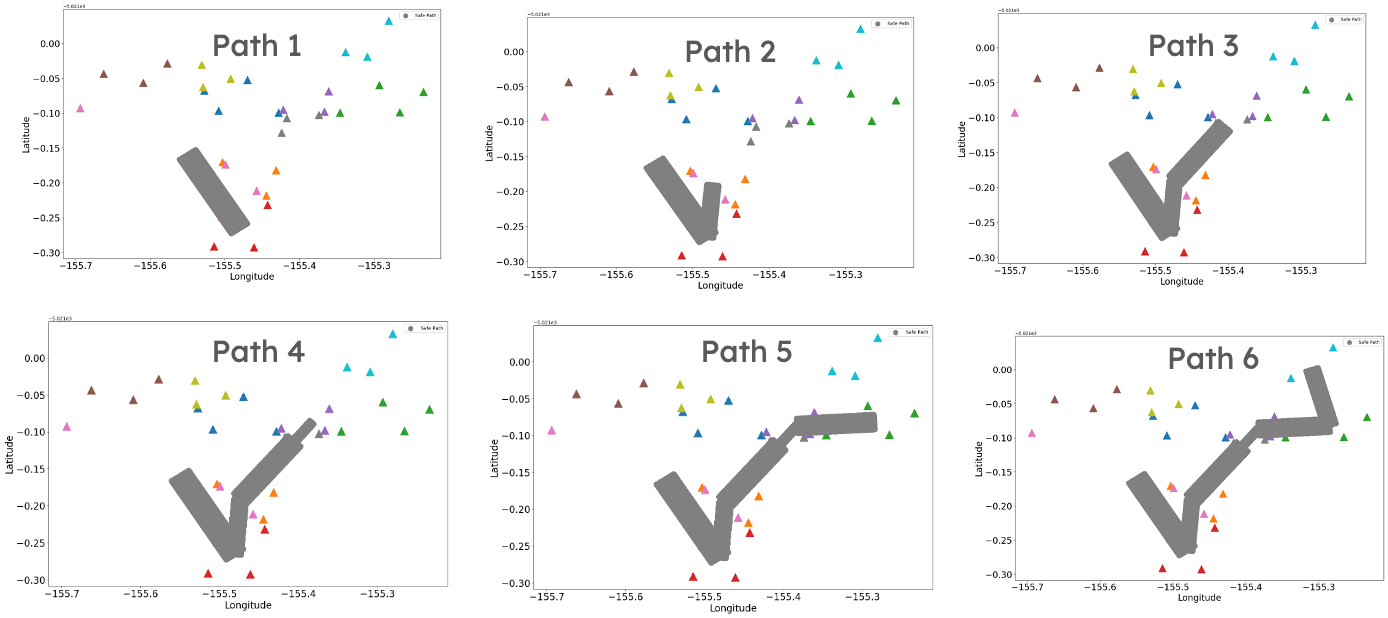}
\centering
\caption{Showing six different safe paths each of different lengths generated from the combination of different clusters in path class 1. } 
\label{path_class1}

\end{figure}

\begin{figure}[!ht]
\includegraphics[width=0.5\textwidth]{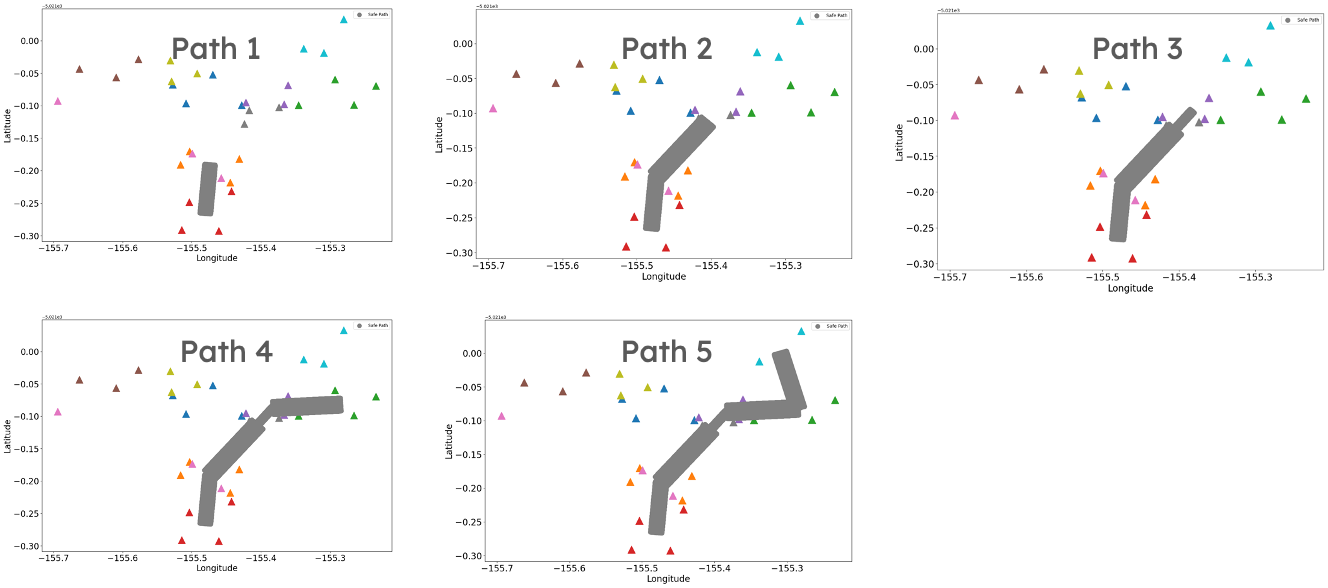}
\centering
\caption{Showing five different safe paths each of different lengths generated from the combination of different clusters in path class 2. } 
\label{path_class3}
\end{figure}

\begin{figure}[!ht]
\includegraphics[width=0.5\textwidth]{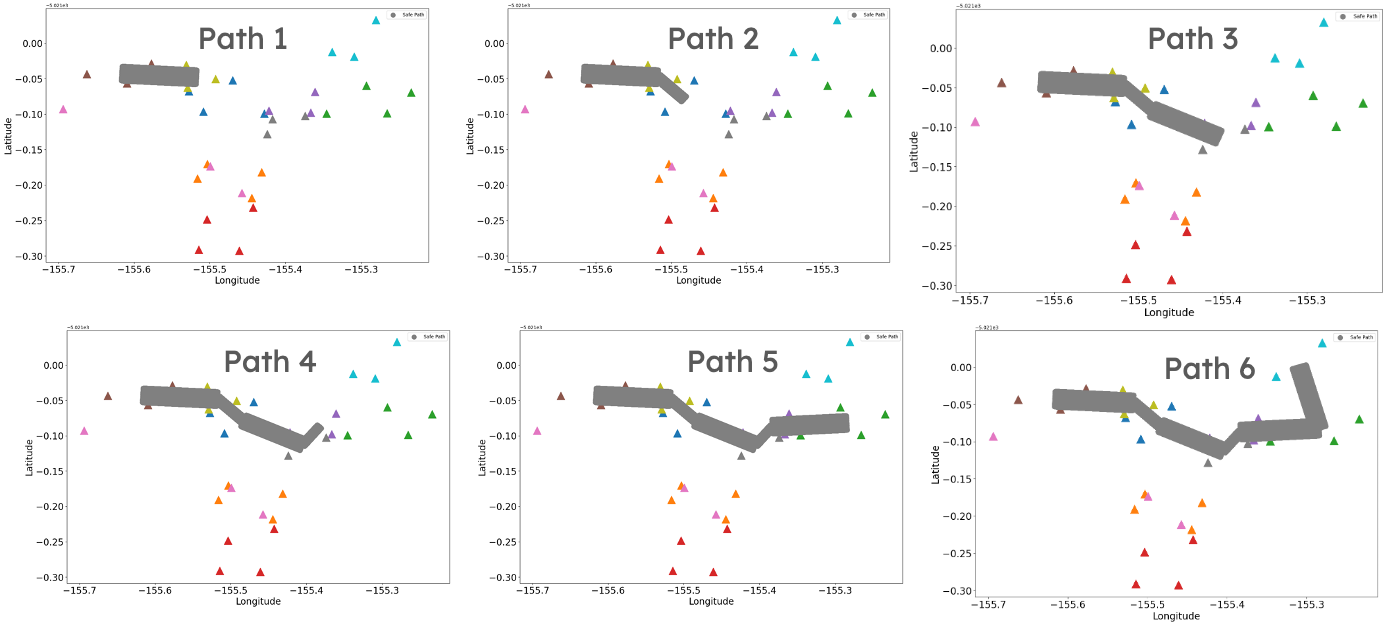}
\centering
\caption{Showing six different safe paths each of different lengths generated from the combination of different clusters in path class 3. } 
\label{path_class2}
\end{figure}

We simulated the 1000 movements of an object along each of the paths belonging to three safe path classes using two different secure navigation approaches. Approach 1 performs the navigation using the battlefield motion model $(BBM)$ for next-state prediction and LanBLoc for positioning, known as SecNav $(BBM +LanBLoc)$. Approach 2 guides the moving object using BBM, EKF, and LanBloc known as SecNav $(BBM +EKF+LanBLoc)$. To test the efficacy of two different approaches, we obtained the trace of the positions that the object traveled along to reach the target zone corresponding to each approach. Safe path classes indicate the group of paths that are drawn from the combination of cluster members associated with that category. The different paths associated with three different safe path classes are shown in Fig. \ref{path_class1}, Fig. \ref{path_class3}, and Fig. \ref{path_class2}. 


We calculated the relative error associated with ground truth trajectory and estimated trajectory and obtained the percent error. We calculated the average displacement error (ADE) and Final displacement error (FDE) of estimated trajectory points and ground truth trajectory points. Results are shown in table \ref{percent_error} and \ref{displacement_errors} below.
\begin{figure}[!ht]
\includegraphics[width=0.5\textwidth]{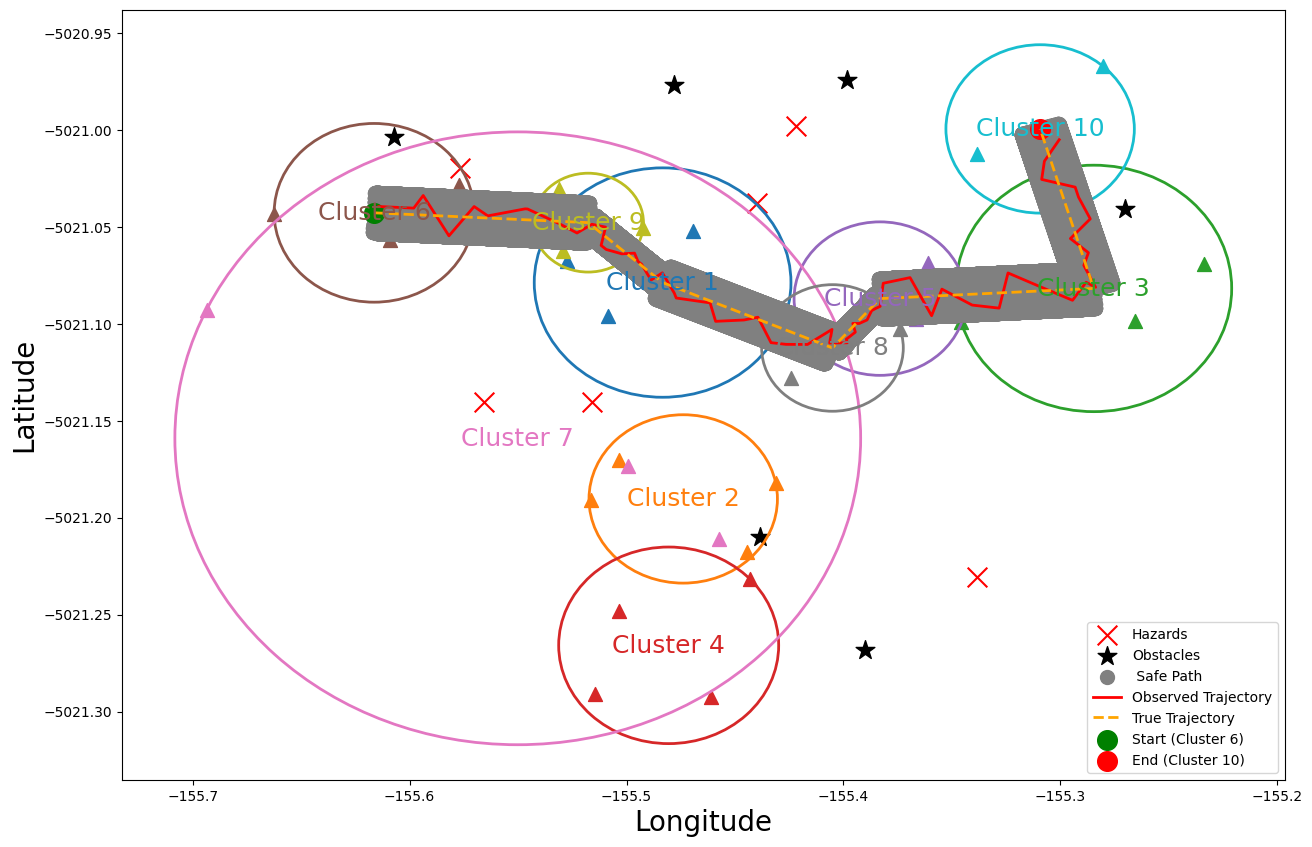}
\centering
\caption{Battlefield map with known landmarks, obstacles, and hazards showing ground truth trajectory and observed trajectory of moving object within the safe path from cluster 6 to cluster 10.} 
\label{safe_path}
\end{figure}
Table \ref{percent_error} shows the percent of error in the length of the estimated and ground truth trajectory for three different path classes shown in Fig.\ref{path_class1},\ref{path_class3},\ref{path_class2}, each containing multiple paths with different lengths. We obtained an Average error of $10.49 \%$  for the SeureNav(BMM+LanbLoc) approach and $6.51\%$ for SecureNav (BMM+EKF+LanBloc). Approach $2$ which uses the Battlefield motion model and LanBloc integrated with the Kalman filter shows an improvement of $4.18\%$,$3.81\%$, and $3.96\%$ in each path class $PC1$, $PC1$, and $PC3$ respectively with an overall improvement of around 4\%.\\ 

\begin{table}[!ht]
\centering 
\begin{tabular}{|l|c|c|c|c|}
\hline
\multirow{2}{*}{Approaches}&\multicolumn{3}{|c|}{ Percent Error }&\multirow{2}{*}{Average}\\\cline {2-4}
& PC1 & PC2 & PC3& \\\cline {2-4} 
\hline {SeureNav(BMM+LanbLoc)} & 9.93 & 11.23 & 10.31 & 10.49 \\
\hline \textbf{SecureNav(BMM+EKF+LanBloc)}& \textbf{5.76} & \textbf{7.42} & \textbf{6.35} &\textbf{6.51} \\
\hline
\end{tabular}
\caption{Comparision of average percent error of trajectories belong to each path class associated with two different approaches.}
\label{percent_error}
\end{table}
Table \ref{displacement_errors} shows the average displacement error and final displacement errors of true trajectories and $1000$ estimated trajectories associated with each safe path. We observed an average $ADE/FDE$ of $5.56/5.99$ for Approach $1$ and $2.97/3.27$ for Approach $2$ which is an improvement of $46.61\%$ on ADE and $45.44\%$ on FDE compared to Approach $1$.
\begin{table}[!ht]
\centering
\begin{tabular}{|l|l|l|l|}
\hline Approaches & Path Class & ADE & FDE \\
\hline
\multirow{4}{*}{SeureNav(BMM+LanbLoc)} & PC1 & 5.13 & 6.26 \\\cline{2-4} 
& PC2 & 5.11 & 5.86 \\ \cline{2-4} & PC3 & 6.45 & 5.86 \\\cline{2-4} 
& Average & 5.56 & 5.99\\
\hline \multirow{4}{*}{\textbf{SecureNav(BMM+EKF+LanBloc)}} & PC1 & 2.73 & 3.16 \\\cline{2-4} 
 & PC2 & 2.72 & 3.08 \\\cline{2-4} 
& PC3 & 3.46 & 3.57 \\\cline{2-4} 
& \textbf{Average} & $\mathbf{2.97}$ & $\mathbf{3 . 2 7}$ \\\cline{2-4}
\hline
\end{tabular}
\caption{Comparision of ADE and FDE of trajectories associated with two different approaches}
\label{displacement_errors}
\end{table}

\section{Conclusion and Future Works}
The proposed Secure Navigation using Landmark-based Localization provides a unique approach for guiding the mobile troops along the predefined safe path in GPS-denied battlefield environments by combining a landmark-based localization $(LanBLoc)$ algorithm and an adapted Extended Kalman filter based on a battlefield motion model. The proposed solution has demonstrated notable accuracy in localization and navigation along defined safe paths. Experimentation and simulation results have validated the effectiveness of this system, showcasing its potential to significantly enhance the safety and operational efficacy of military forces in challenging environments. This approach not only addresses the critical vulnerability of GPS reliance but also introduces a robust framework for real-time, adaptive navigation strategy, ensuring troops can maneuver safely and efficiently in contested or hazardous areas. Our future research will focus on refining path-planning algorithms for more efficient navigation and developing lightweight localization algorithms for real-time positioning that would significantly boost the capability of military forces to navigate safely and effectively in challenging or hostile environments.

\end{document}